\documentclass[10pt,twocolumn,letterpaper]{article}

\usepackage{wacv}
\usepackage{fancyhdr}
\usepackage{times}
\usepackage{epsfig}
\usepackage{graphicx}
\usepackage{amsmath}
\usepackage{amssymb}
\usepackage{booktabs}
\usepackage{tabularray}
\usepackage{pifont}
\usepackage{makecell}
\usepackage{float}
\usepackage[table,xcdraw]{xcolor}
\usepackage{caption}
\usepackage{subcaption}

\newcommand\doubleplus{+\kern-1.3ex+\kern0.8ex}
\newcommand\mdoubleplus{\ensuremath{\mathbin{+\mkern-10mu+}}}
\newcommand{\cmark}{\ding{51}}%
\newcommand{\xmark}{\ding{55}}%

\DeclareMathOperator*{\concat}{\mdoubleplus}

\fancypagestyle{firstpage}
{
    \fancyhead[R]{}    
    \fancyhead[L]{This is a preprint of a paper published at the WACV 2023 workshop on real world surveillance. Please cite: \textit{Klemen Grm, Berk Kemal Özata, Vitomir Štruc, Hazım Kemal Ekenel. Meet-in-the-middle: Multi-scale upsampling and matching \\ for cross-resolution face recognition. WACV workshops, 2023.} }
}

\def\wacvPaperID{36} 

\wacvapplicationstrack 

\wacvfinalcopy 

\ifwacvfinal
\usepackage[breaklinks=true,bookmarks=false]{hyperref}
\else
\usepackage[pagebackref=true,breaklinks=true,colorlinks,bookmarks=false]{hyperref}
\fi

\begin{document}

\title{Meet-in-the-middle: Multi-scale upsampling and matching \\ for cross-resolution face recognition }

\author{Klemen Grm\textsuperscript{1} \qquad Berk Kemal Özata\textsuperscript{2,3} \qquad Vitomir Štruc\textsuperscript{1} \qquad Hazım Kemal Ekenel\textsuperscript{2}\\
\textsuperscript{1}University of Ljubljana, \textsuperscript{2}Istanbul Technical University, \textsuperscript{3}ASELSAN Inc.\\
{\tt\small \{klemen.grm, vitomir.struc\}@fe.uni-lj.si, \{ozata19, ekenel\}@itu.edu.tr}}


\maketitle
\thispagestyle{empty}
\thispagestyle{firstpage}

\begin{abstract}
   In this paper, we aim to address the large domain gap between high-resolution face images, e.g., from professional portrait photography, and low-quality surveillance images, e.g., from security cameras. Establishing an identity match between disparate sources like this is a classical surveillance face identification scenario, which continues to be a challenging problem for modern face recognition techniques. To that end, we propose a method that combines face super-resolution, resolution matching, and multi-scale template accumulation to reliably recognize faces from long-range surveillance footage, including from low quality sources. The proposed approach does not require training or fine-tuning on the target dataset of real surveillance images. Extensive experiments show that our proposed method is able to outperform even existing methods fine-tuned to the SCFace dataset. 
   
\end{abstract}

\section{Introduction}

Recent advances in deep learning methods, including model architectures, loss functions, datasets, and training procedures have enabled considerable improvements in the performance of face recognition systems. However, real-life face identification in challenging conditions, such as surveillance scenarios, remain an open challenge. 

Despite common claims that super-resolution and face hallucination can aid in face recognition performance in this regime, systematic bias studies have found them to be poorly applicable to real-life scenarios due to the domain gap between real and synthetically downsampled low-resolution images~\cite{grm2019face,wang2021unsupervised}.

In this paper, we present a framework for addressing this issue, and we make the following novel contributions:

\begin{itemize}
    \item We develop a face hallucination method that reliably upsamples low-resolution probe images at multiple scales.
    \item We develop a multi-scale face recognition model that improves recognition performance by matching the resolution of the probe and gallery images being compared.
    \item We develop a face template derivation method that reduces the effects of image quality by accumulating the features of a given image over multiple scales.
    \item The proposed approach is independent of employed deep face recognition model. 
    That is, it can be combined with any deep face model.
    Moreover, it does not require any fine-tuning on the target dataset.
\end{itemize}

The combination of our contributions results in a strongly performing face recognition method that achieves state-of-the-art rank-1 identification rates (IR) on the challenging Surveillance Cameras Face Database (SCFace,~\cite{grgic2011scface}).

\begin{figure*}[]
    \centering
    \includegraphics[width=0.93\linewidth]{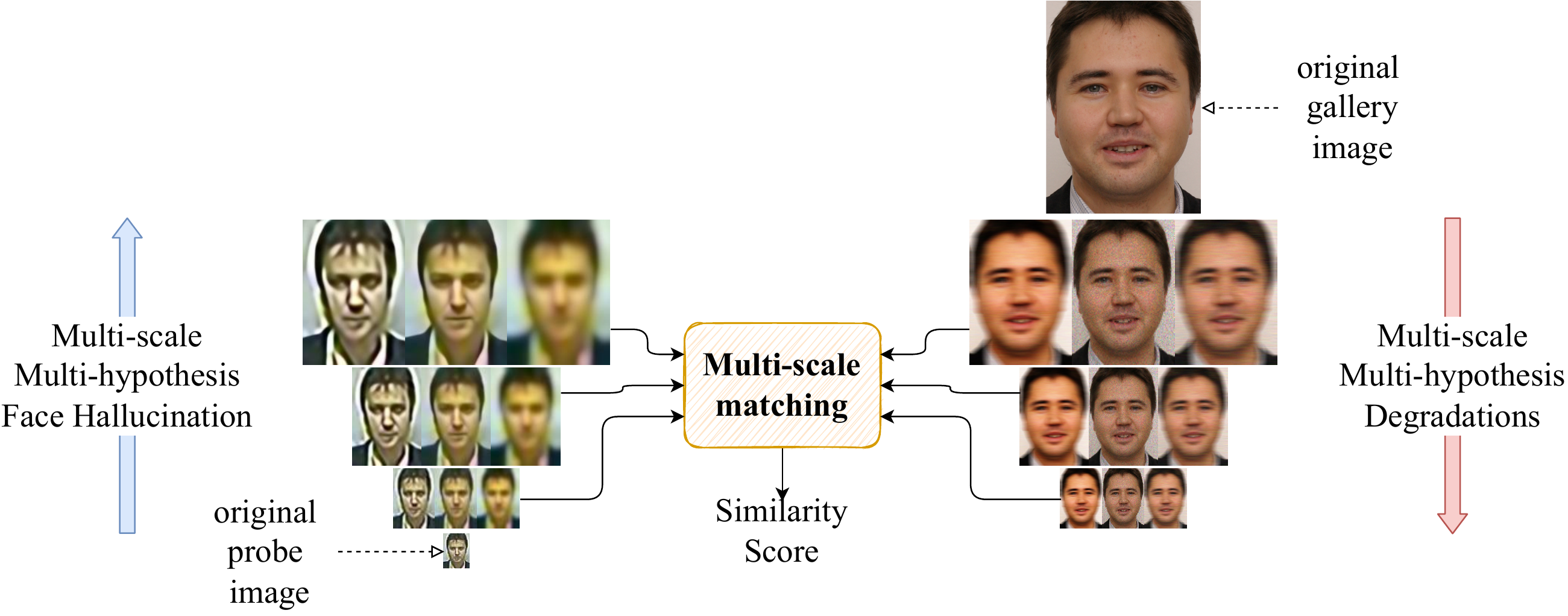}
    \caption{Overall diagram of the proposed method. The inputs to the method are a high-resolution gallery image, and a low-quality probe image. Our method re-samples both to a series of intermediate scales, and generate multiple hypotheses at every scale. Our multi-scale matching method then makes use of all the images generated by this process to improve face verification performance.
    }\label{fig:cross-res-diagram}
\end{figure*}

\section{Related work}

\textbf{Super-resolution.} Recently, a significant amount of research has gone into applying modern deep learning techniques to the problem of super-resolution. Supervised learning approaches typically work by generating a dataset of low-resolution and high-resolution image pairs, for example, by using a dataset of high-resolution images as the ground-truth (target), and subjecting each image to a pre-defined degradation process to derive the training inputs. A model, e.g., a Convolutional Neural Network (CNN) or similar deep neural network, is then trained to upsample the input image, using gradient descent over a pixel reconstruction error metric, such as mean square error or mean absolute error~\cite{dong2015image,kim2016accurate,lim2017enhanced}. 

More recent approaches to general super-resolution include more complex loss functions, such as perceptual loss~\cite{johnson2016perceptual}. In this case, a secondary \textit{loss network} is used as the criterion instead of directly measuring pixel errors. Alternatively, in adversarial approaches, the high-resolution output is rated by a discriminator, a model trained to discern between real and generated high-resolution images~\cite{ledig2017photo,wang2018enhanced,wang2021real}. This approach allows super-resolution models to traverse the trade-off between reconstruction fidelity and realistic appearance in a principled manner~\cite{blau2018perception}.

\textbf{Face hallucination.} General super-resolution methods, i.e., methods capable of upsampling images of arbitrary subjects, are strictly limited in their reconstruction capability by the amount of information present in the input image. Face hallucination refers to super-resolution methods applied specifically to the domain of human faces. By limiting the domain of the training data and the application domain, these methods are able to achieve useful reconstructions at higher magnification factors, up to $8\times$ the resolution of the input image, than is typical for general super-resolution methods, i.e., up to $4\times$~\cite{zhang2018super,grm2020face,chen2020learning,lu2021face}. 

\textbf{Face recognition.} Modern work on large-scale face recognition has consisted of parallel developments in collecting large face databases and training large-scale deep learning models to efficiently incorporate them into face template derivation models~\cite{sun2014deep,BMVC2015_41,cao2018vggface2,msceleb}. These models are typically trained using some combination of classification and metric learning loss functions. More recently, research work has been focused on developing novel loss functions that combine the benefits of classification and metric learning~\cite{wang2018cosface,deng2019arcface} and explicitly account for the quality of the input image~\cite{kim2022adaface}.

\textbf{Cross-resolution face recognition.}
The cross-resolution face recognition approaches can be categorized into three main groups: resolution-invariant methods \cite{lu2018deep, khalid2020resolution, huang2020improving, massoli2020cross, amato2020multi}, face hallucination based methods \cite{zhang2018super, yin2020fan}, and degradation based methods \cite{aghdam2019exploring, fang2020generate}.

Resolution-invariant methods try to minimise the difference between the representations of low-resolution (LR) and high-resolution (HR) face images.
Lu et al. \cite{lu2018deep} proposed the Deep Coupled ResNet (DCR) model consisting one trunk network and two branch networks. 
They first train the trunk network with face images of different resolutions, then two branch networks are trained to learn coupled-mappings between LR and HR face images. 
Knowledge distillation based models \cite{khalid2020resolution, huang2020improving, massoli2020cross, amato2020multi} distill the information from a \emph{Teacher} network, which is pre-trained with high-resolution face images, to the \emph{Student} network, which is trained on images of different resolutions.

Face hallucination based methods reconstruct high-resolution face images from low-resolution ones and they perform recognition in the HR domain.
In \cite{zhang2018super}, identity preserving face hallucination method is proposed.
It utilizes super-identity loss that penalizes the identity difference between high-resolution and super-resolved face images. 
Feature Adaptation Network (FAN) \cite{yin2020fan} disentangles the features into identity and non-identity components and performs face normalization while improving the resolution. 

In contrast to the face hallucination based methods, degradation based methods transform high-resolution faces to low-resolution ones. 
In \cite{aghdam2019exploring}, it is shown that a simple resolution matching technique that downsamples HR gallery images to the resolution of LR probe images improves the cross-resolution face recognition performance. 
Another approach employs a GAN based method, Resolution Adaption Network \cite{fang2020generate}, that realistically transforms HR images into the LR domain and uses a feature adaption network to extract LR information from HR embedding.


\section{Methodology}

To solve the cross-resolution face recognition problem, we propose a method that will make the quality of low-resolution and high-resolution images converge in the image domain.
Specifically, we increase the quality of the probe images using face hallucination networks and degrade the gallery images by applying combinations of degradation functions.
We propose a multi-hypothesis approach for face hallucination that is robust to artefacts caused by image degradations.
We also make use of multi-scale representations of faces by applying different magnification factors.
On the other hand, multiple degraded hypotheses are generated from each gallery and resolution-matched to the multi-scale probe hypotheses.
Finally, we use multi-scale template accumulation in order to measure the similarity between a pair of probe and gallery image.
The overall method is depicted in Figure \ref{fig:cross-res-diagram}.

\subsection{Face template extraction}

We use state-of-the-art pretrained face recognition models to extract templates from degraded gallery images and super-resolved probe images. We compare different template networks, namely, the ResNet~\cite{he2015deep} family of models (ResNet-50, ResNet101) trained using the ArcFace~\cite{deng2019arcface} loss function.
The networks are trained on the MS1M~\cite{msceleb} and Glint360k~\cite{an2021partial} datasets, which contain 8M and 17M face images, respectively.

\subsection{Gallery degradations}
\label{sec:gallery-degradations}
In order to decrease the domain gap between low-resolution probe face images and high-resolution gallery face images, we reduce the quality of the gallery images by applying multiple types of degradation functions at different scales.
Multiple hypotheses for each gallery image are obtained by applying every possible combination of the degradations.
The length of the combination determines the number of downsampling that occur between each degradation of that combination.
We set the maximum length of the combinations to three for this process.

For example, given the gallery image $G$ and given the combination $c_i =\ \  <\rho^{i_1}, \rho^{i_2}, \rho^{i_3}>$ consisting of degradation functions $\rho^{i_j} \in \left\{\rho_1, \rho_2, ..., \rho_l\right\}$, where $j \in \left\{1, 2, 3\right\}$, the $i$-th degraded hypothesis $G_{H_i}$ is obtained using the following equation:
\begin{equation}
G_{H_i} = \rho^{i_3}\left(\downarrow_s\left(\rho^{i_2}\left(\downarrow_s\left(\rho^{i_1}\left(G\right)\right)\right)\right)\right),
\end{equation}
where $\downarrow_s$ denotes the downsampling operation with a scale factor of two and performed by the bicubic function.
The degradation function types $\left\{\rho_1, \rho_2, ..., \rho_l\right\}$ are given in Table \ref{table:degradation-set}.

After obtaining the degraded gallery face hypotheses, we also match the resolution of each hypothesis to the resolution of the probe image of interest. 
Note that, for the experiments with multi-scale probe images, we obtain multiple resolution-matched images from each degraded gallery image for each probe scale factor.

\begin{figure*}[]
    \centering
    \includegraphics[width=0.5\linewidth]{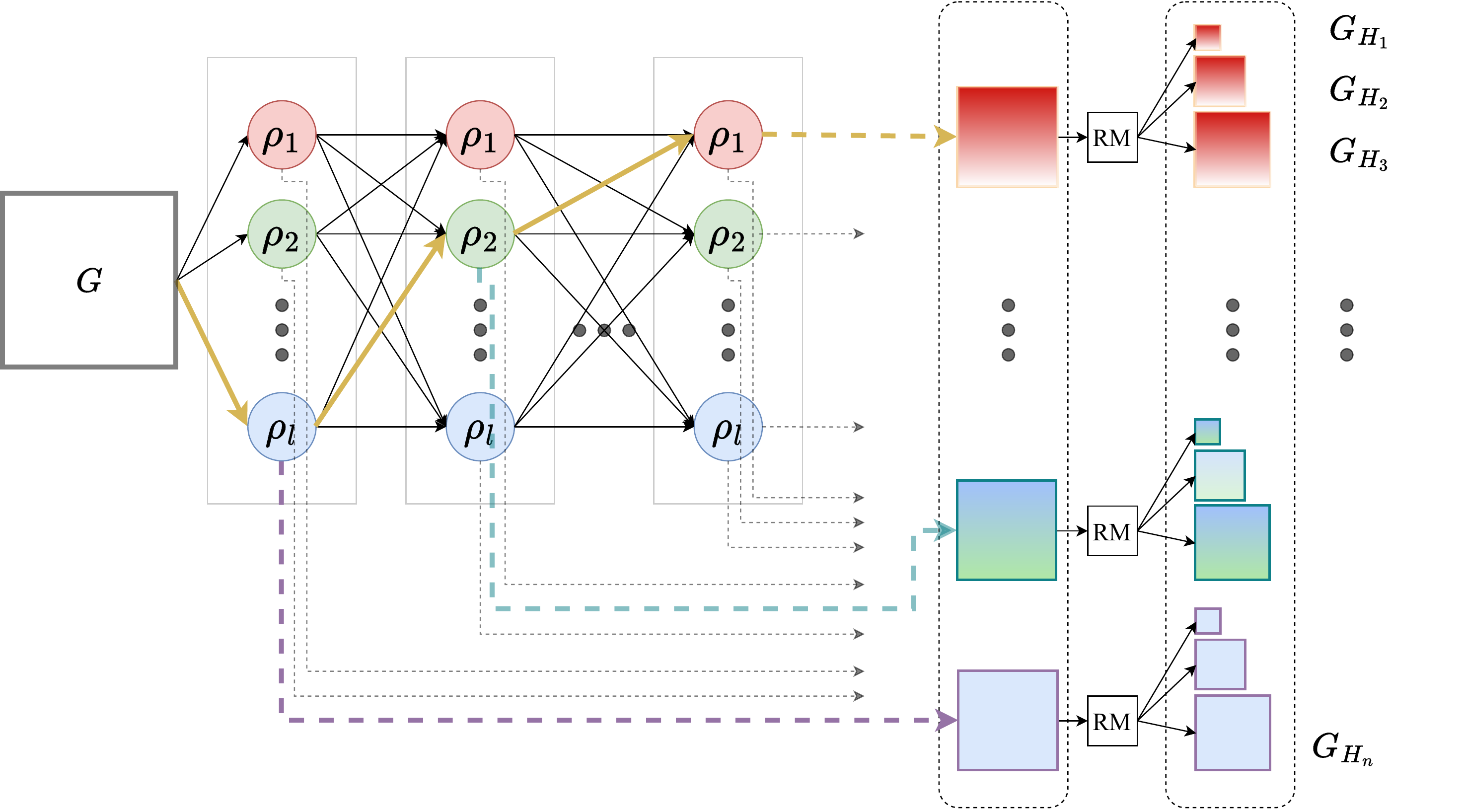}
    \caption{Gallery degradation process overview.
    In the graph above, there are $l$ different degradation options to be applied at each step.  
    All the possible paths in the graph generate a degraded gallery image and all of them are used in the recognition pipeline. 
    Note that, downsampling operation is applied between each degradation.
    We also match the resolution of the degraded gallery images with the resolution of the probe images, which is done at the block denoted with RM (resolution-matching).
    For example, the combination highlighted with yellow lines has three degradations to be applied sequentially.
    On the other hand, blue and purple dashed lines show the particular combinations of length 1 and 2, respectively.
    }\label{fig:degradation-diagram}
\end{figure*}

\begin{table}[]
\begin{center}
\scalebox{0.9}{
\begin{tblr}{
    colspec     = {X[1,l]X[2,l]}
}
\hline[1pt,solid]
Degradation Function   &   Description \\
\hline
 Additive Gaussian noise     & Additive Gaussian noise with mean 0 and variance 0.02 \\
 Speckle noise               & Multiplicative Gaussian noise is added to the image with mean 0 and variance 0.02 \\
 Color jitter                & Randomly change hue and saturation \\
 Brightness jitter           & Randomly change brightness and contrast \\
 Motion blur                 & Horizontal motion blur is applied with a window size 20 \\
 Gaussian blur               & Gaussian blur with sigma 1.1 and window size 5 \\
 Disk blur                   & Disk blur with radius 5 \\
 Perspective transform       & Random perspective transform \\
 Shear mapping               & Random shear transformation \\
 Upscale following downscale & First downscale the image, then upscale back to the original resolution \\
 Patch shuffle               & Random patch shuffle \cite{patch-shuffle} \\
\hline[1pt,solid]
\end{tblr}
}
\end{center}
\caption{List of degradation functions}
\label{table:degradation-set}
\end{table}

\begin{figure}[]
    \centering
    \includegraphics[width=0.8\linewidth]{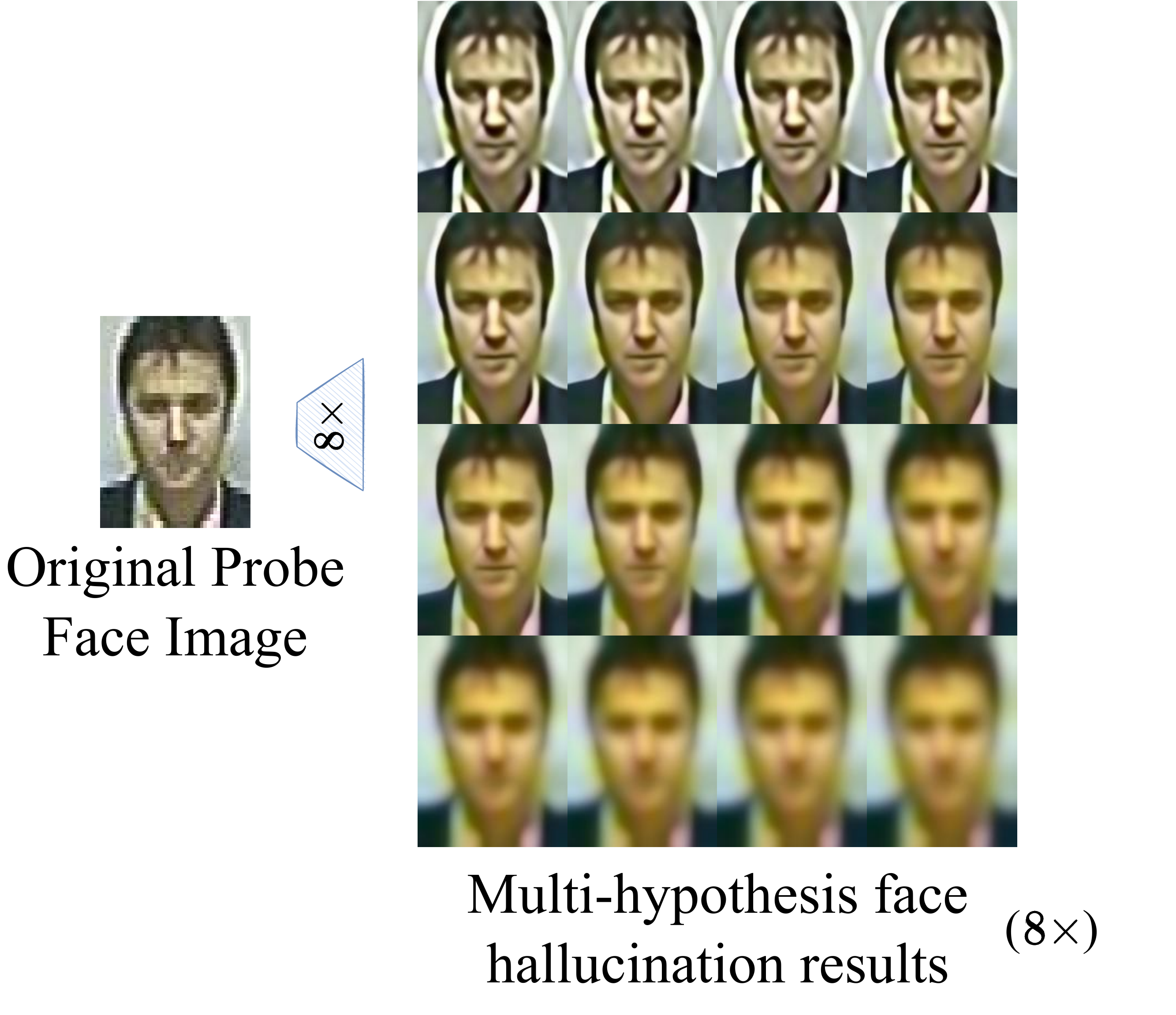}
    \caption{Multi-hypothesis face hallucination results
    }\label{fig:multi-hypothesis-face-hallucination}
\end{figure}

\subsection{Face hallucination}
\label{sec:face-hallucination}
In order to add high-resolution details to real life low-resolution surveillance images, we train a variant of the EDSR~\cite{lim2017enhanced} super-resolution convolutional neural network exclusively on face images. By limiting the training set to face images, as opposed to general computer vision datasets such as, ImageNet~\cite{imagenet} or DIV2K~\cite{agustsson2017ntire} typically used for super-resolution training, the network is able to learn to upsample human faces in more detail, which enables a higher magnification factor ($8\times$) than is typically used for general super-resolution methods (up to $4\times$).
Our super-resolution network is trained on the VGGFace2~\cite{cao2018vggface2} dataset with 3M images. 
The reference images represent the ground-truth, and the training inputs are derived by applying a degradation (downsampling) pipeline to the full-resolution images. 

Given a super-resolution model $f_{SR}$ capable of upsampling an input low-resolution image $x$, i.e., such that $y = f_{SR}(x)$ where $x\in\mathbb{R}^{h\times w\times 3}$, $y\in\mathbb{R}^{kh\times kw\times 3}$, and $k$ is the upsampling factor, we adapt a \textit{multi-hypothesis} upsampling approach. We notice that many low-resolution images are corrupted beyond their limited spatial resolution, e.g., by noise and sampling artefacts. In order to alleviate these artefacts, we blur the input image to different extents. Specifically, we generate 16 versions of the input images by blurring them using a Gaussian kernel with $\sigma=0$ through $\sigma=1$. Each of the images is then upsampled separately using our super-resolution model. We present examples of the multi-hypothesis super-resolution in Figure \ref{fig:multi-hypothesis-face-hallucination}. We note that the unblurred low-resolution image results in a suboptimal reconstruction, since the super-resolution model amplifies its noise and artefacts to an extent. Similarly, if the input image is blurred too much, this results in a blurry reconstruction.

To determine the best hypothesis to use out of the generated super-resolved images, we extract a face recognition template from each of the hypothesis images, and pick the one with the best similarity score to any of the images in the gallery set.

In the rest of the sections, we discuss the means to use the outputs of this multi-hypothesis model in order to improve face recognition from these images.

\begin{figure*}[]
    \centering
    \includegraphics[width=1.0\linewidth]{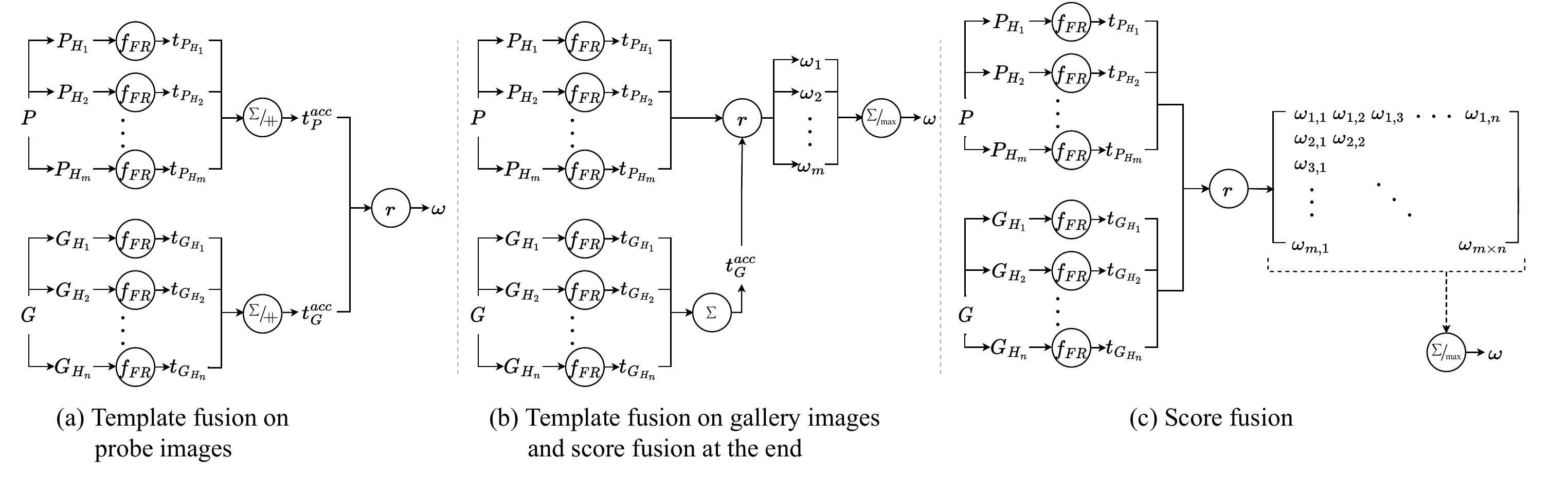}
    \caption{
    Different combinations of fusion methods while comparing a pair of probe image and gallery image. We start with obtaining hypotheses for the probe image $P$ and the gallery image $G$. Then, a face template is obtained from each hypothesis image using face recognition model $f_{FR}$. Finally, template and/or score fusion methods are applied to obtain a single similarity score. 
    }\label{fig:fusion}
\end{figure*}

\subsection{Multi-scale template accumulation}

\label{sec:multi-scale-template-accumulation}

Next, we consider how to optimally combine our capability to degrade the gallery images to arbitrary resolutions, and our family of face super-resolution models that can magnify input images by factors of $2\times$, $4\times$ and $8\times$. 
In this step, we measure the similarity between each probe and gallery image pair.
Then, the identification is performed by nearest-neighbor classifier.

We use face recognition (template extraction) models $f_{FR}$, which take a face image as an input and generate a face template vector as an output. 
We assume that the templates extracted from each of the scales, $f_{FR}(x_{s_i}) \forall s_i\in \left\{48px, 96px, 192px\right\}$, include useful information about the identity of the subjects, as well as noise related to the resolution and quality of the image, i.e.,

\begin{equation}
f_{FR}(x_{s_i}) = I_x + \epsilon_{s_i},
\label{eqn:resolution_noise}
\end{equation}

where $x_{s_i}$ is an input image at scale $s_i$, $I_x$ is the useful information about the subject of the image $x$, and $\epsilon_{s_i}$ is the template noise arising from the unwanted effects of image resolution and quality.  Under the assumption that $\epsilon_{s_i}$ is normally distributed for all scales, i.e., that $\epsilon_{s_i} \sim \mathcal{N}\left(0, \mathbf{\Sigma} \right)\forall s_i\in \left\{48px, 96px, 192px \right\}$, we try to reduce the impact of the noise term by accumulating the three scale representations to derive the final templates. Under the above assumption, we expect this to suppress the resolution-related noise while preserving the identity information.


Given the probe image $P$ and the gallery image $G$, we obtain the multi-hypothesis super resolved images $\left\{P_{H_1}, P_{H_2}, ..., P_{H_{m \times |s|}}\right\}$ and the degraded and resolution-matched gallery images $\left\{G_{H_1}, G_{H_2}, ..., G_{H_{n \times |s|}}\right\}$, where $m$ is the number of super-resolution hypotheses for each scale, $n$ is the number of degraded images, and $|s|$ is the number of scales. 
Using a face recognition model $f_{FR}$, we then extract a face template from each of the images, such that $\left\{t_{P_{H_i}}=f_{FR}\left( P_{H_i} \right)\right\}$ and $\left\{t_{G_{H_j}}=f_{FR}\left( G_{H_j} \right)\right\}$.


To compare two given face templates, we calculate their similarity scores using the correlation distance metric:
\begin{equation}
r(t_{P_{H_i}}, t_{G_{H_j}}) = \frac{(t_{P_{H_i}} - \overline{t_{P_{H_i}}})\cdot (t_{G_{H_j}} - \overline{t_{G_{H_j}}})^\mathsf{T}}{{||(t_{P_{H_i}} - \overline{t_{P_{H_i}}})||}_2{||(t_{G_{H_j}} - \overline{t_{G_{H_j}}})||}_2},
\end{equation}
where $t_{P_{H_i}}$ is the template of the $i$-th probe hypothesis and $t_{G_{H_j}}$ is the template of the $j$-th gallery hypothesis being compared, and $\overline{t_{P_{H_i}}}$ and $\overline{t_{G_{H_j}}}$ are the mean probe and gallery templates, respectively.

We use different strategies to compare the templates of a probe image and the templates of a gallery image. 
These strategies include template fusion and similarity score fusion.
We also use a hybrid approach in which both the template fusion and the similarity score fusion are applied.

\textbf{Template fusion.}
The first fusion strategy is to obtain a single template for the probe image and a single template for the gallery image, before calculating the similarity score between them.
By doing so, we reduce the number of hypotheses to be compared to a single hypothesis.
In order to do that, we use two types of fusion methods, namely, template addition and template concatenation.
The template addition is element-wise summation of the templates.
Given multi-scale probe templates $\left\{t_{P_{H_i}} \in \mathbb{R}^D\right\}$ and multi-scale gallery templates $\left\{t_{G_{H_j}} \in \mathbb{R}^D\right\}$, where $D$ is the dimensionality of the template vector, the template addition is performed as:
\begin{equation}
\label{eq:template-addition}
t_P^{acc} = \sum_i{t_{P_{H_i}}}, \qquad t_G^{acc} = \sum_j{t_{G_{H_j}}},
\end{equation}
where $t_P^{acc} \in \mathbb{R}^D$ and $t_G^{acc} \in \mathbb{R}^D$ are the accumulated face templates for the probe image and the gallery image, respectively.

The template concatenation requires some care before applying because the dimensions should match after the concatenation operation for similarity calculation. We first accumulate the templates of each scale separately using the template addition to obtain $\left\{t_{P_{s_x}}\right\}$ and $\left\{t_{G_{s_x}}\right\}$, where $s_x$ refers to the $x\times$ magnification scale with regards to the low resolution image.
Then, we concatenate the feature vectors obtained at each scale:
\begin{equation}
t_P^{acc} = \concat_x\left({t_{P_{s_x}}}\right), \qquad t_G^{acc} = \concat_x\left({t_{G_{s_x}}}\right),
\end{equation}
where $\concat$ is the concatenation operator and joins the face templates derived from individual scale images into a single vector. Here, $t_P^{acc} \in \mathbb{R}^{D|s|}$ and $t_G^{acc} \in \mathbb{R}^{D|s|}$ are the accumulated face templates for the probe image and the gallery image, where $|s|$ is the number of scales.

\textbf{Similarity score fusion.} In this strategy, we do not fuse the face templates, instead, we calculate every possible similarity score between the face templates of a probe image and the face templates of a gallery image.
Then, we fuse these similarity scores by either adding them up as proposed in~\cite{kittler1998combining} or by defining the similarity between the given underlying gallery and probe image as the maximal similarity between any pair of hypotheses generated.

The different combinations of the fusion methods are illustrated in Figure \ref{fig:fusion}. 
Here, the correlation similarity metric $r$ measures the similarity scores between each possible probe and gallery template pairs. 
The template addition in equation (\ref{eq:template-addition}) can be applied to both probe hypotheses and gallery hypotheses, or it can be applied to only one of them, as shown in Figure \ref{fig:fusion}(b). 
However, while using template concatenation, it should be applied to both probe hypotheses and gallery hypotheses, since the dimensions should be compatible for similarity measurement.

In Figure \ref{fig:fusion}(c), we first obtain the similarity scores between each pair of probe templates and gallery templates, resulting in a $m \times n$ similarity score matrix.
In this similarity matrix, $i$-th row contains the similarity of the $i$-th probe hypothesis with each gallery hypothesis.
Next, we first select a score fusion method and use that method to calculate row-wise score.
Then, we choose another score fusion method and use it to obtain a single similarity value from the similarity score vector produced in the previous 
step.


\section{Experiments}

\begin{figure}[b]
    \centering
    \includegraphics[width=0.8\linewidth]{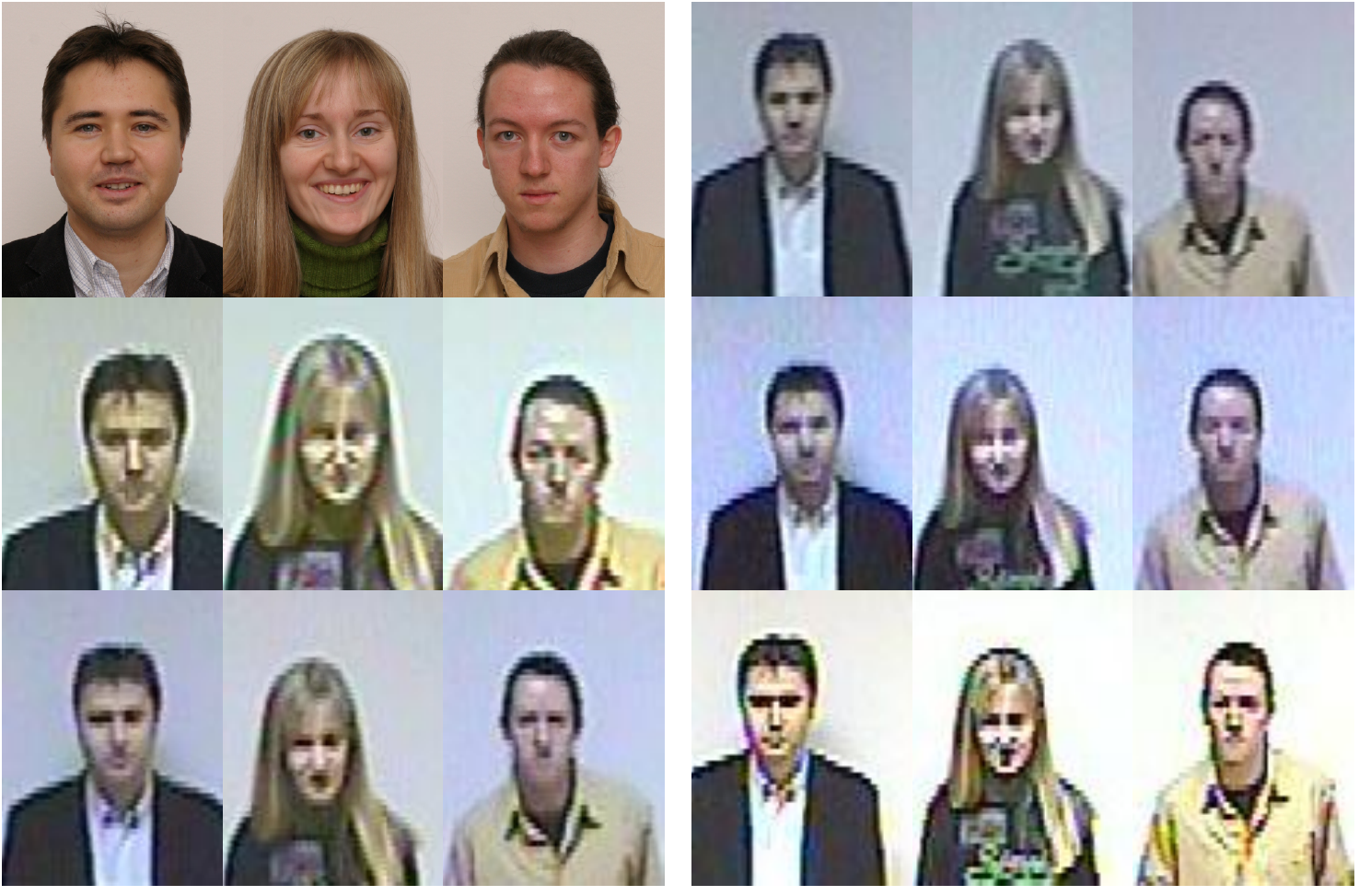}
    \caption{Sample images from SCFace dataset. Gallery images (first three images of the first column) and probe images from five different cameras at $4.2m$ distance. 
    }\label{fig:sample-images-scface}
\end{figure}

\begin{table}[]
\begin{center}
\scalebox{0.85}{
\begin{tabular}{@{}llrrr@{}}
\toprule
\textbf{Model}                    &                         & \multicolumn{3}{c}{\textbf{Rank-1 IR (\%)}}                                    \\ \midrule
\rowcolor[HTML]{EFEFEF} 
\textit{Baseline}                 &                         &                         &                         &                            \\ \midrule
MS1MV3-R50                        &                         &                         & \multicolumn{1}{l}{}    & 39.84                      \\
MS1MV3-R101                       &                         &                         & \multicolumn{1}{l}{}    & 50.61                      \\
Glint360k-R50                     &                         &                         & \multicolumn{1}{l}{}    & 64.00                      \\
Glint360k-R101                    &                         &                         & \multicolumn{1}{l}{}    & 74.61                      \\ \midrule
\rowcolor[HTML]{EFEFEF} 
\textit{+Single Hypothesis SR}    &                         & $2\times$               & $4\times$               & $8\times$                  \\ \midrule
MS1MV3-R50                        &                         & 41.23                   & 32.92                   & 27.69                      \\
MS1MV3-R101                       &                         & 50.15                   & 41.07                   & 36.92                      \\
Glint360k-R50                     &                         & 67.07                   & 55.07                   & 47.53                      \\
Glint360k-R101                    &                         & 73.07                   & 60.92                   & 52.92                      \\ \midrule
\rowcolor[HTML]{EFEFEF} 
\textit{+Multi Hypothesis SR}     &                         & $2\times$               & $4\times$               & $8\times$                  \\ \midrule
MS1MV3-R50                        &                         & 50.92                   & 53.07                   & 52.61                      \\
MS1MV3-R101                       &                         & 56.00                   & 58.46                   & 60.15                      \\
Glint360k-R50                     &                         & 67.84                   & 74.15                   & 72.00                      \\
Glint360k-R101                    &                         & 74.15                   & 76.61                   & 76.30                      \\ \midrule
\rowcolor[HTML]{EFEFEF} 
\textit{+Resolution Matching}     &                         & $2\times$               & $4\times$               & $8\times$                  \\ \midrule
MS1MV3-R50                        &                         & 60.61                   & 54.92                   & 50.92                      \\
MS1MV3-R101                       &                         & 70.00                   & 62.00                   & 57.53                      \\
Glint360k-R50                     &                         & 83.53                   & 74.76                   & 70.92                      \\
Glint360k-R101                    &                         & 86.61                   & 80.00                   & 76.61                      \\ \midrule
\rowcolor[HTML]{EFEFEF} 
\textit{+Gallery Degradations}    &                         & $2\times$               & $4\times$               & $8\times$                  \\ \midrule
MS1MV3-R50                        &                         & 70.61                   & 68.61                   & 65.53                      \\
MS1MV3-R101                       &                         & 79.38                   & 77.84                   & 76.30                      \\
Glint360k-R50                     &                         & 84.61                   & 83.84                   & 84.61                      \\
Glint360k-R101                    &                         & 90.00                   & 89.69                   & 88.00                      \\ \midrule
\rowcolor[HTML]{EFEFEF} 
\textit{+Multi-scale Accumulation} & $\text{S}_{\text{max}}$ & $\text{S}_{\text{add}}$ & $\text{T}_{\text{add}}$ & $\text{T}_{\text{concat}}$ \\ \midrule
MS1MV3-R50                        & 78.46                   & 72.00                   & 70.61                   & 71.69                      \\
MS1MV3-R101                       & 78.92                   & 75.84                   & 79.38                   & 79.23                      \\
Glint360k-R50                     & 86.76                   & 88.15                   & 88.61                   & 89.23                      \\
Glint360k-R101                    & 90.15                   & 90.76                   & 90.15                   & 91.07                      \\ \bottomrule
\end{tabular}
}
\end{center}
\caption{
Ablation studies on the d1 distance of the SCFace dataset. 
$2 \times$, $4 \times$, and $8 \times$ correspond to the magnification factors applied to the probe faces.
$\text{S}_{\text{max}}$, $\text{S}_{\text{add}}$, $\text{T}_{\text{add}}$, and $\text{T}_{\text{concat}}$ correspond to finding the maximum score, score addition, template addition and template concatenation operations, respectively.
}
\label{table:scface-d1-ablation-study}
\end{table}

\subsection{Dataset}
We use the SCFace dataset~\cite{grgic2011scface} for the experiments, which contain 130 subjects.
The \emph{gallery set} consists of one high-quality, high resolution, frontal image of each of the subjects.
The \emph{probe set} consists of a series of low resolution images of the same subjects. Probe face images are captured using five different surveillance cameras and from three different distances, namely, $4.2m$, $2.6m$ and $1m$. Sample images from the dataset are shown in Figure~\ref{fig:sample-images-scface}. As modern face recognition methods are capable of achieving close to $100\%$ rank-1 recognition accuracy on the closer $1m$ and $2.6m$ probe sets, we focus our experiments on the $4.2m$ probe image set.
Faces are detected using MTCNN~\cite{mtcnn}, then cropped by enlarging the bounding boxes with a scale factor of 1.3 following the findings in~\cite{aghdam2019exploring}.
The average resolution of the cropped faces is $22\times 29$ for $\text{d1} = 4.2m$.

Ablation experiments are conducted on the entire dataset, whereas for comparison with the previous work 10 Repeated Random Sub-Sampling Validation (RRSSV) is performed for randomly selected 80 subjects.

\subsection{Ablation study}
\label{sec:ablation_study}

In order to determine the contributions of the individual components of our proposed method, 
we conduct an extensive ablation study on the SCFace dataset.
The results are given in Table \ref{table:scface-d1-ablation-study}. 
In the baseline method, we simply use only the original probe faces and gallery faces, and resize them to the input size of the face recognition model $f_{FR}$. 

Next, we add the components of our proposed method one by one. 
First, we obtain single hypothesis super-resolved probe faces using the face hallucination network with the magnification factors of $2\times$, $4\times$, $8\times$. There is a slight improvement when the magnification factor is $2 \times$. However, the performance reduces when the scale factor is increased, as can be seen from the results of the experiments with $4 \times$ and $8 \times$ scale factors.  This may be due to high-frequency artefacts caused by real-world image degradations being amplified at high magnification factors.

Then, we examined the effect of using multi hypothesis face hallucination method instead of single-hypothesis method. In this method, we try to minimize the artefacts by super-resolving probe images at different blur levels and obtain multiple hypotheses for each probe image.
We then find the maximum similarity between the templates obtained from a given probe image's hypotheses and the template of a given gallery image.
After that, the recognition is done by assigning the subject whose gallery image has the highest similarity score.
In all of the scale factors, the performance improves against both the baseline and the single-hypothesis face hallucination method.

So far, we have only worked on the probe face images.
With resolution matching, we match the size of the gallery face images to the size of the probe faces.
Then, they are resized to the input size of the face recognition model.
The results of the experiments with $2 \times$ and $4 \times$ scale factors show that the resolution matching helps to cross-resolution matching process. 
However, since the probe face sizes at scale factor $8 \times$ ($192px$) are larger than the network input size ($112px$), we do not see any improvement in experiments with this scale factor.
The best results were obtained when the scale factor is taken $2 \times$, where the gallery images have been degraded most compared to other scales. 

By applying different types of degradation combinations, we aim to model real-world image degradations. 
This will reduce the image quality difference between probe and gallery images.
Using the method described in \mbox{section \ref{sec:gallery-degradations}}, we generate multiple hypotheses for each gallery face image.
Then, we compare the multi-hypothesis probe templates and multi-hypothesis gallery templates as illustrated in \mbox{Figure \ref{fig:fusion}(c)} using the maximum score fusion.
The recognition performance is increased at all scales when gallery degradations are used.

Finally, by adding multi-scale template accumulation, we obtain the entire proposed face recognition pipeline.
Here, we try to benefit from multi-scale representations of the probe images and gallery images.
In order to do that, we use different types of fusion methods to accumulate the information at different scales ($2 \times$, $4 \times$, and $8 \times$).
As a general trend, performance has improved in all fusion methods.
We also examined the combinations of different fusion methods in \mbox{Table \ref{table:scface-d1-ablation-study-on-fusion-methods}}. 
In the first part of the table, we fix the template addition $\text{T}_{\text{add}}$ on probe images and afterwards combine it with different fusion methods.
There is no significant difference between the fusion methods $\text{T}_{\text{add}}$ (on gallery images) and score addition $\text{S}_{\text{add}}$, which were later combined.
However, using the max rule ($\text{S}_{\text{max}}$) improves the performance.
In this case, we have a single accumulated template for the probe image and the score fusion is performed with respect to multiple gallery hypotheses.
We see that selecting the most similar gallery hypotheses is better than accumulating them. 
In the second part of the \mbox{Table \ref{table:scface-d1-ablation-study-on-fusion-methods}}, this time, we fix the template addition $\text{T}_{\text{add}}$ on gallery images.
In this case, there is no significant difference between any of the combined fusion methods.

We can apply $S_{max}$ and $S_{add}$ sequentially as described in section \ref{sec:multi-scale-template-accumulation}.
Following the findings from the ablation studies, we first apply $S_{max}$ on the similarity matrix in order to obtain the most similar gallery hypotheses, then $S_{add}$ is performed to obtain a single similarity score.
By applying $S_{max}$ and $S_{add}$ respectively with a ResNet-101 based face recognition model trained on Glint360k dataset, we achieve \mbox{93.53\%} Rank-1 IR on the d1 distance of the SCFace dataset.

\begin{table}[]
\begin{center}
\scalebox{0.85}{
\begin{tabular}{llrrr}
\toprule
model                                                     &         & \multicolumn{1}{l}{$\text{T}_{\text{add}}$} & \multicolumn{1}{l}{$\text{S}_{\text{add}}$} & \multicolumn{1}{l}{$\text{S}_{\text{max}}$} \\ \midrule
\multicolumn{2}{l}{\textit{Fix $\text{T}_{\text{add}}$ on the probe templates}} & \multicolumn{1}{l}{}                        & \multicolumn{1}{l}{}                        & \multicolumn{1}{l}{}                        \\ \midrule
MS1MV3-R50                                                &         & 70.61                                       & 70.30                                       & 78.15                                       \\
MS1MV3-R101                                               &         & 79.38                                       & 78.61                                       & 85.53                                       \\
Glint360k-R50                                             &         & 88.61                                       & 88.15                                       & 92.46                                       \\
Glint360k-R101                                            &         & 90.15                                       & 91.07                                       & 92.00                                       \\ \midrule
\textit{Fix $\text{T}_{\text{add}}$ on the gallery templates}         &         & \multicolumn{1}{l}{}                        & \multicolumn{1}{l}{}                        & \multicolumn{1}{l}{}                        \\ \midrule
MS1MV3-R50                                                &         & 70.61                                       & 71.23                                       & 78.46                                       \\
MS1MV3-R101                                              &         & 79.38                                       & 77.07                                       & 78.92                                       \\
Glint360k-R50                                              &         & 88.61                                       & 89.07                                       & 86.76                                       \\
Glint360k-R101                                           &         & 90.15                                       & 90.76                                       & 90.15                                       \\ \bottomrule
\end{tabular}
}
\end{center}
\caption{
Ablation study on different fusion methods on the d1 distance of the SCFace dataset. 
Rank-1 IR (\%) results are given. 
In the first part of the table, templates of the probe hypotheses are fused with template addition $\text{T}_{\text{add}}$ and followed by different types of fusion methods as shown in the columns. 
In the second part of the table, the templates of the gallery hypotheses are first fused with template addition $\text{T}_{\text{add}}$, unlike the first part.
}
\label{table:scface-d1-ablation-study-on-fusion-methods}
\end{table}

\begin{table}[]
\begin{center}
\scalebox{0.9}{
\begin{tabular}{lcr}
\toprule
Model & Fine-Tuning & d1 (\%)  \\
\midrule
Martinez et al. \cite{martinez2021benchmarking}      &   \xmark       &       68.3          \\
Fang et al. \cite{fang2020generate}                  &   \xmark       &       70.5          \\
Aghdam et al. \cite{aghdam2019exploring}                &   \xmark       &       78.5          \\
Lai et al. \cite{lai2021deep}                        &   \xmark       &       79.7          \\
Khalid et al. \cite{safwan2021npt}                   &   \xmark       &       85.7          \\
Khalid et al. \cite{khalid2020resolution}            &   \xmark       &  \textbf{88.3}      \\
\midrule         
Sun et al. \cite{sun2020classifier}                  &   \cmark       &       65.5          \\
DCR \cite{lu2018deep}                                &   \cmark       &       73.3          \\
TCN \cite{zha2019tcn}                                &   \cmark       &       74.6          \\
FAN \cite{yin2020fan}                                &   \cmark       &       77.5          \\
Fang et al. \cite{fang2020generate}                  &   \cmark       &       81.3          \\
Huang et al. \cite{huang2020improving}               &   \cmark       &       86.8          \\
Li et al. \cite{li2022deep}                          &   \cmark       &       90.4          \\
Lai et al. \cite{lai2021deep}                        &   \cmark       &       93.0          \\
Martinez et al. \cite{martinez2021benchmarking}       &   \cmark       &  \textbf{95.3}      \\
\midrule          
Ours                                                 &   \xmark       &  \textbf{95.4}      \\
\bottomrule
\end{tabular}
}
\end{center}
\caption{Comparison of Rank-1 IR (\%) results on the SCFace dataset with previous works.
Models fine-tuned on the SCFace dataset are denoted with a checkmark on \emph{Fine-Tuning} column.
The average of 10 RRSSV for 80 subjects out of 130 subjects is reported for our models.
}
\label{table:scface-sota}
\end{table}



\subsection{Face template comparison}

We use four different face template models ($f_{FR}$) in the above experiments, namely, ResNet-50 and ResNet-101 networks trained on either MS1MV3 or Glint360k datasets.

From the results in Table~\ref{table:scface-d1-ablation-study}, we notice that as a general trend, face templates derived using ResNet-101 based models tend to perform better than the ones derived using the ResNet-50 based models. This implies that the model capacity is not yet saturated on either of these large-scale training datasets, and recognition performance could benefit from larger CNN models.

It is also clear that all else being equal, using templates from the models trained on the larger Glint360k datasets is always preferable to the ones trained on MS1MV3. This shows that larger datasets and more expressive deep models continue to be the main driver of advances in face recognition performance. In addition, the distribution of images over the dataset subjects in Glint360k is closer to previously established heuristics regarding the optimal number of subjects given a fixed dataset size~\cite{bansal2017dos}.

\subsection{Comparison with previous works}

In Table \ref{table:scface-sota}, we compare our results on the SCFace dataset with the ones from the previous works. 
To be comparable with the previous works, in this section, we report the mean of 10 Repeated Random Sub-Sampling Validation (RRSSV) result for randomly selected 80 subjects.
These results are obtained using ResNet-101 model trained on Glint360k dataset.
We applied all of the proposed blocks -- multi-hypothesis face hallucination, resolution matching, gallery degradations, and multi-scale accumulation.
The multi-scale accumulation is implemented by applying $S_{max}$ and $S_{add}$ to the gallery and probe templates, respectively.

Methods that perform fine-tuning on randomly selected 50 subjects are specified in Table \ref{table:scface-sota} with a check mark on \emph{Fine-Tuning} column.
Our model achieves the highest performance with 95.4\% accuracy for distance d1$=4.2m$ over randomly selected 80 subjects.
This model achieves \mbox{93.53\%} accuracy over all 130 subjects of the SCFace dataset.
The next best result among the methods that do not use the SCFace dataset for training purposes, 88.3\%, is achieved with \cite{khalid2020resolution}.
These results are obtained without performing any training or fine-tuning on the target dataset, yet performed close to or even better than the ones that utilize a part of the dataset for training purposes.

\section{Conclusion}

\begin{figure}[]
    \centering
    \includegraphics[width=0.95\linewidth]{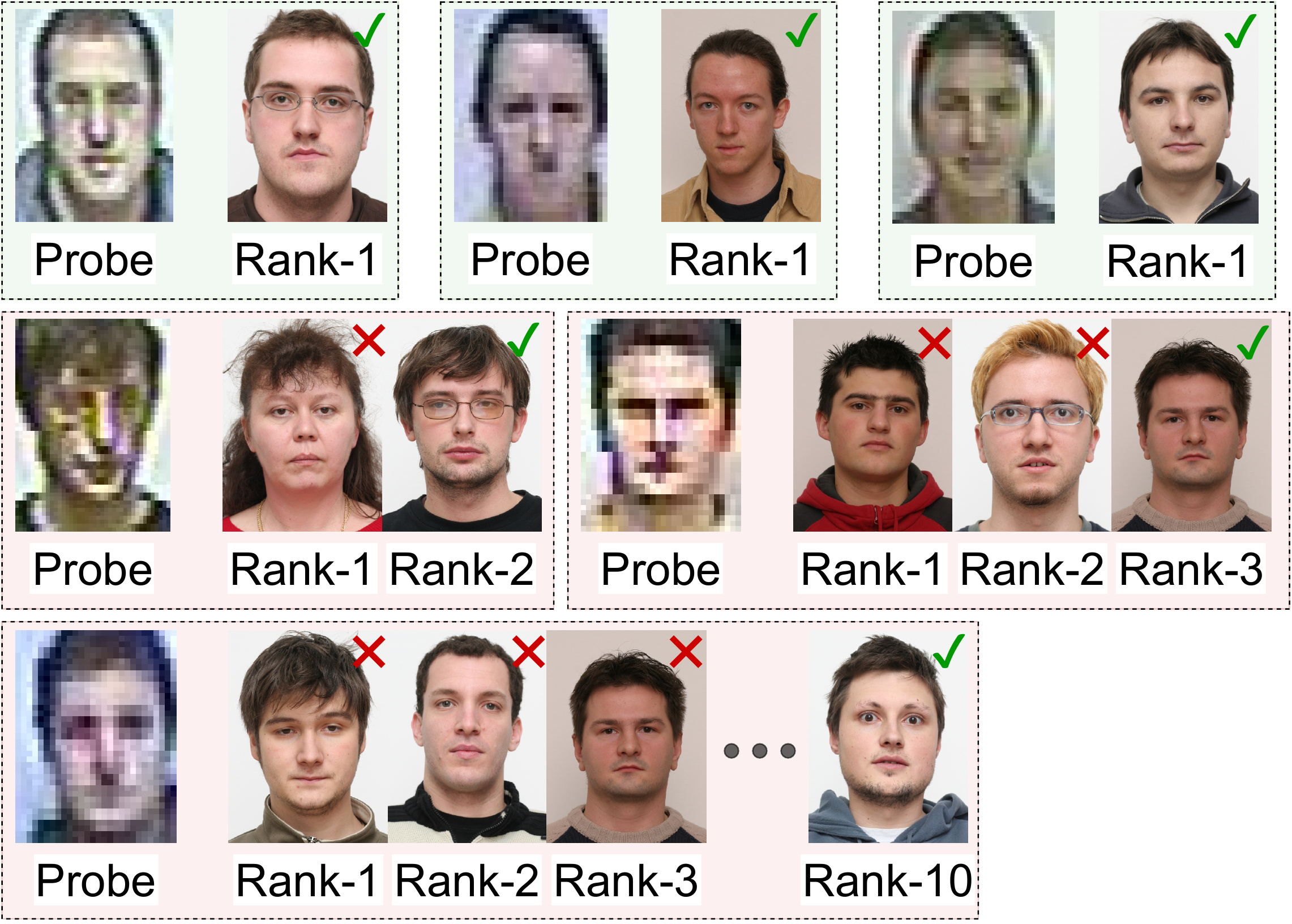}
    \caption{Samples of correctly identified probe images (top row) and failure cases (rest of the figure). The failure cases are ordered by the identification rank of the correct gallery image.
    }\label{fig:failure-cases}
\end{figure}


We have presented a novel method that enables reliable cross-resolution face recognition. 
The experimental results show that each of the components of our proposed method, including the face hallucination, gallery degradation, and multi-scale matching, contributes towards our final result.

Given the high rank-1 accuracy of our method, we are interested in studying the individual failure cases on the SCFace dataset. We present some failure cases in Figure \ref{fig:failure-cases}. We note that the 
failure cases in the SCFace dataset mostly correspond to very low-quality probe images where a large amount of ambiguity as to the identity of the subjects exists. Even then, the proposed method matches it to a sensible identity, e.g., one that matches the gender, haircut, or hair colour of the probe image - the only attributes that can be reliably determined from the failure cases presented.

\section*{Acknowledgement}
This research was supported by the bilateral Slovenian Research Agency (ARRS) and the Scientific and Technological Research Council of Türkiye (TUBITAK) funded project: Low Resolution Face Recognition (FaceLQ), with TUBITAK project number 120N011.

{\small
\bibliographystyle{ieee_fullname}
\bibliography{egbib}

\begin{thebibliography}{10}\itemsep=-1pt

\bibitem{aghdam2019exploring}
Omid Abdollahi~Aghdam, Behzad Bozorgtabar, Hazim Kemal~Ekenel, and
  Jean-Philippe Thiran.
\newblock Exploring factors for improving low resolution face recognition.
\newblock In {\em Proceedings of the IEEE/CVF Conference on Computer Vision and
  Pattern Recognition (CVPR) Workshops}, June 2019.

\bibitem{agustsson2017ntire}
Eirikur Agustsson and Radu Timofte.
\newblock Ntire 2017 challenge on single image super-resolution: Dataset and
  study.
\newblock In {\em Proceedings of the IEEE conference on computer vision and
  pattern recognition workshops}, pages 126--135, 2017.

\bibitem{amato2020multi}
G. Amato, F. Falchi, C. Gennaro, F.~V. Massoli, and C. Vairo.
\newblock Multi-resolution face recognition with drones.
\newblock In {\em SSIP}, 2020.

\bibitem{an2021partial}
Xiang An, Xuhan Zhu, Yuan Gao, Yang Xiao, Yongle Zhao, Ziyong Feng, Lan Wu, Bin
  Qin, Ming Zhang, Debing Zhang, and Ying Fu.
\newblock Partial fc: Training 10 million identities on a single machine.
\newblock In {\em Proceedings of the IEEE/CVF International Conference on
  Computer Vision (ICCV) Workshops}, pages 1445--1449, October 2021.

\bibitem{bansal2017dos}
Ankan Bansal, Carlos Castillo, Rajeev Ranjan, and Rama Chellappa.
\newblock The dos and donts for cnn-based face verification.
\newblock In {\em Proceedings of the IEEE International Conference on Computer
  Vision Workshops}, pages 2545--2554, 2017.

\bibitem{blau2018perception}
Yochai Blau and Tomer Michaeli.
\newblock The perception-distortion tradeoff.
\newblock In {\em Proceedings of the IEEE conference on computer vision and
  pattern recognition}, pages 6228--6237, 2018.

\bibitem{cao2018vggface2}
Qiong Cao, Li Shen, Weidi Xie, Omkar~M Parkhi, and Andrew Zisserman.
\newblock Vggface2: A dataset for recognising faces across pose and age.
\newblock In {\em 2018 13th IEEE international conference on automatic face \&
  gesture recognition (FG 2018)}, pages 67--74. IEEE, 2018.

\bibitem{chen2020learning}
Chaofeng Chen, Dihong Gong, Hao Wang, Zhifeng Li, and Kwan-Yee~K Wong.
\newblock Learning spatial attention for face super-resolution.
\newblock {\em IEEE Transactions on Image Processing}, 30:1219--1231, 2020.

\bibitem{imagenet}
Jia Deng, Wei Dong, Richard Socher, Li-Jia Li, Kai Li, and Li Fei-Fei.
\newblock Imagenet: A large-scale hierarchical image database.
\newblock In {\em 2009 IEEE Conference on Computer Vision and Pattern
  Recognition}, pages 248--255, 2009.

\bibitem{deng2019arcface}
Jiankang Deng, Jia Guo, Niannan Xue, and Stefanos Zafeiriou.
\newblock Arcface: Additive angular margin loss for deep face recognition.
\newblock In {\em Proceedings of the IEEE/CVF Conference on Computer Vision and
  Pattern Recognition (CVPR)}, June 2019.

\bibitem{dong2015image}
Chao Dong, Chen~Change Loy, Kaiming He, and Xiaoou Tang.
\newblock Image super-resolution using deep convolutional networks.
\newblock {\em IEEE transactions on pattern analysis and machine intelligence},
  38(2):295--307, 2015.

\bibitem{fang2020generate}
H. Fang, W. Deng, Y. Zhong, and J. Hu.
\newblock Generate to adapt: Resolution adaption network for surveillance face
  recognition.
\newblock In {\em ECCV}, 2020.

\bibitem{grgic2011scface}
Mislav Grgic, Kresimir Delac, and Sonja Grgic.
\newblock Scface--surveillance cameras face database.
\newblock {\em Multimedia tools and applications}, 51(3):863--879, 2011.

\bibitem{grm2019face}
Klemen Grm, Martin Pernuš, Leo Cluzel, Walter~J. Scheirer, Simon Dobrišek,
  and Vitomir Štruc.
\newblock Face hallucination revisited: An exploratory study on dataset bias.
\newblock In {\em Proceedings of the IEEE/CVF Conference on Computer Vision and
  Pattern Recognition (CVPR) Workshops}, June 2019.

\bibitem{grm2020face}
Klemen Grm, Walter~J. Scheirer, and Vitomir Štruc.
\newblock Face hallucination using cascaded super-resolution and identity
  priors.
\newblock {\em IEEE Transactions on Image Processing}, 29:2150--2165, 2020.

\bibitem{msceleb}
Yandong Guo, Lei Zhang, Yuxiao Hu, Xiaodong He, and Jianfeng Gao.
\newblock Ms-celeb-1m: A dataset and benchmark for large-scale face
  recognition.
\newblock In {\em European conference on computer vision}, pages 87--102.
  Springer, 2016.

\bibitem{he2015deep}
Kaiming He, Xiangyu Zhang, Shaoqing Ren, and Jian Sun.
\newblock Deep residual learning for image recognition.
\newblock {\em arXiv preprint arXiv:1512.03385}, 2015.

\bibitem{huang2020improving}
Y. Huang, P. Shen, Y. Tai, S. Li, X. Liu, J. Li, F. Huang, and R. Ji.
\newblock Improving face recognition from hard samples via distribution
  distillation loss.
\newblock In {\em ECCV}, 2020.

\bibitem{johnson2016perceptual}
Justin Johnson, Alexandre Alahi, and Li Fei-Fei.
\newblock Perceptual losses for real-time style transfer and super-resolution.
\newblock In {\em European conference on computer vision}, pages 694--711.
  Springer, 2016.

\bibitem{patch-shuffle}
G. Kang, X. Dong, L. Zheng, and Y. Yang.
\newblock Patchshuffle regularization.
\newblock {\em arXiv preprint arXiv:1707.07103}, 2017.

\bibitem{khalid2020resolution}
Syed~Safwan Khalid, Muhammad Awais, Zhen-Hua Feng, Chi-Ho Chan, Ammarah Farooq,
  Ali Akbari, and Josef Kittler.
\newblock Resolution invariant face recognition using a distillation approach.
\newblock {\em IEEE Transactions on Biometrics, Behavior, and Identity
  Science}, 2(4):410--420, 2020.

\bibitem{kim2016accurate}
Jiwon Kim, Jung~Kwon Lee, and Kyoung~Mu Lee.
\newblock Accurate image super-resolution using very deep convolutional
  networks.
\newblock In {\em Proceedings of the IEEE conference on computer vision and
  pattern recognition}, pages 1646--1654, 2016.

\bibitem{kim2022adaface}
Minchul Kim, Anil~K. Jain, and Xiaoming Liu.
\newblock Adaface: Quality adaptive margin for face recognition.
\newblock In {\em Proceedings of the IEEE/CVF Conference on Computer Vision and
  Pattern Recognition (CVPR)}, pages 18750--18759, June 2022.

\bibitem{kittler1998combining}
Josef Kittler, Mohamad Hatef, Robert~PW Duin, and Jiri Matas.
\newblock On combining classifiers.
\newblock {\em IEEE transactions on pattern analysis and machine intelligence},
  20(3):226--239, 1998.

\bibitem{lai2021deep}
Shun-Cheung Lai and Kin-Man Lam.
\newblock Deep siamese network for low-resolution face recognition.
\newblock In {\em 2021 Asia-Pacific Signal and Information Processing
  Association Annual Summit and Conference (APSIPA ASC)}, pages 1444--1449.
  IEEE, 2021.

\bibitem{ledig2017photo}
Christian Ledig, Lucas Theis, Ferenc Husz{\'a}r, Jose Caballero, Andrew
  Cunningham, Alejandro Acosta, Andrew Aitken, Alykhan Tejani, Johannes Totz,
  Zehan Wang, et~al.
\newblock Photo-realistic single image super-resolution using a generative
  adversarial network.
\newblock In {\em Proceedings of the IEEE conference on computer vision and
  pattern recognition}, pages 4681--4690, 2017.

\bibitem{li2022deep}
Peiying Li, Shikui Tu, and Lei Xu.
\newblock Deep rival penalized competitive learning for low-resolution face
  recognition.
\newblock {\em Neural Networks}, 2022.

\bibitem{lim2017enhanced}
Bee Lim, Sanghyun Son, Heewon Kim, Seungjun Nah, and Kyoung Mu~Lee.
\newblock Enhanced deep residual networks for single image super-resolution.
\newblock In {\em Proceedings of the IEEE conference on computer vision and
  pattern recognition workshops}, pages 136--144, 2017.

\bibitem{lu2021face}
Tao Lu, Yuanzhi Wang, Yanduo Zhang, Yu Wang, Liu Wei, Zhongyuan Wang, and
  Junjun Jiang.
\newblock Face hallucination via split-attention in split-attention network.
\newblock In {\em Proceedings of the 29th ACM International Conference on
  Multimedia}, pages 5501--5509, 2021.

\bibitem{lu2018deep}
Z. Lu, X. Jiang, and A. Kot.
\newblock Deep coupled resnet for low-resolution face recognition.
\newblock {\em IEEE Signal Processing Letters}, 2018.

\bibitem{martinez2021benchmarking}
Yoanna Martinez-Diaz, Miguel Nicolas-Diaz, Heydi Mendez-Vazquez, Luis~S
  Luevano, Leonardo Chang, Miguel Gonzalez-Mendoza, and Luis~Enrique Sucar.
\newblock Benchmarking lightweight face architectures on specific face
  recognition scenarios.
\newblock {\em Artificial Intelligence Review}, 54(8):6201--6244, 2021.

\bibitem{massoli2020cross}
F.~V. Massoli, G. Amato, and F. Falchi.
\newblock Cross-resolution learning for face recognition.
\newblock {\em Image and Vision Computing}, 2020.

\bibitem{BMVC2015_41}
Omkar~M. Parkhi, Andrea Vedaldi, and Andrew Zisserman.
\newblock Deep face recognition.
\newblock In {\em Proceedings of the British Machine Vision Conference (BMVC)},
  pages 41.1--41.12, September 2015.

\bibitem{safwan2021npt}
Syed Safwan~Khalid, Muhammad Awais, Chi-Ho Chan, Zhenhua Feng, Ammarah Farooq,
  Ali Akbari, and Josef Kittler.
\newblock Npt-loss: A metric loss with implicit mining for face recognition.
\newblock {\em arXiv e-prints}, pages arXiv--2103, 2021.

\bibitem{sun2020classifier}
Jingna Sun, Yehu Shen, Wenming Yang, and Qingmin Liao.
\newblock Classifier shared deep network with multi-hierarchy loss for low
  resolution face recognition.
\newblock {\em Signal Processing: Image Communication}, 82:115766, 2020.

\bibitem{sun2014deep}
Yi Sun, Yuheng Chen, Xiaogang Wang, and Xiaoou Tang.
\newblock Deep learning face representation by joint
  identification-verification.
\newblock In {\em Advances in Neural Information Processing Systems}, pages
  1988--1996, 2014.

\bibitem{wang2018cosface}
Hao Wang, Yitong Wang, Zheng Zhou, Xing Ji, Dihong Gong, Jingchao Zhou, Zhifeng
  Li, and Wei Liu.
\newblock Cosface: Large margin cosine loss for deep face recognition.
\newblock In {\em Proceedings of the IEEE Conference on Computer Vision and
  Pattern Recognition (CVPR)}, June 2018.

\bibitem{wang2021unsupervised}
Wei Wang, Haochen Zhang, Zehuan Yuan, and Changhu Wang.
\newblock Unsupervised real-world super-resolution: A domain adaptation
  perspective.
\newblock In {\em Proceedings of the IEEE/CVF International Conference on
  Computer Vision (ICCV)}, pages 4318--4327, October 2021.

\bibitem{wang2021real}
Xintao Wang, Liangbin Xie, Chao Dong, and Ying Shan.
\newblock Real-esrgan: Training real-world blind super-resolution with pure
  synthetic data.
\newblock In {\em Proceedings of the IEEE/CVF International Conference on
  Computer Vision (ICCV) Workshops}, pages 1905--1914, October 2021.

\bibitem{wang2018enhanced}
Xintao Wang, Ke Yu, Shixiang Wu, Jinjin Gu, Yihao Liu, Chao Dong, Yu Qiao, and
  Chen Change~Loy.
\newblock Esrgan: Enhanced super-resolution generative adversarial networks.
\newblock In {\em Proceedings of the European Conference on Computer Vision
  (ECCV) Workshops}, September 2018.

\bibitem{yin2020fan}
X. Yin, Y. Tai, Y. Huang, and X. Liu.
\newblock Fan: Feature adaptation network for surveillance face recognition and
  normalization.
\newblock In {\em ACCV}, 2020.

\bibitem{zha2019tcn}
Juan Zha and Hongyang Chao.
\newblock Tcn: Transferable coupled network for cross-resolution face
  recognition.
\newblock In {\em ICASSP 2019-2019 IEEE International Conference on Acoustics,
  Speech and Signal Processing (ICASSP)}, pages 3302--3306. IEEE, 2019.

\bibitem{zhang2018super}
Kaipeng Zhang, Zhanpeng Zhang, Chia-Wen Cheng, Winston~H Hsu, Yu Qiao, Wei Liu,
  and Tong Zhang.
\newblock Super-identity convolutional neural network for face hallucination.
\newblock In {\em Proceedings of the European conference on computer vision
  (ECCV)}, pages 183--198, 2018.

\bibitem{mtcnn}
K. Zhang, Z. Zhang, Z. Li, and Y. Qiao.
\newblock Joint face detection and alignment using multitask cascaded
  convolutional networks.
\newblock {\em IEEE Signal Processing Letters}, 23(10):1499--1503, 2016.

\end{thebibliography}
}

\end{document}